\documentclass[journal]{IEEEtran}
\usepackage{graphicx}
\usepackage[utf8]{inputenc}
\usepackage{subfigure}
\usepackage{gensymb}
\usepackage[noadjust]{cite}
\usepackage{amsmath, amssymb}
\usepackage{color}

\definecolor{darkgray}{rgb}{0.5, 0.5, 0.5}

\DeclareMathOperator*{\argmax}{arg\,max}

\newcommand{\dd}{\mathrm{d}}

\begin{document}
\title{Regularized directional representations for medical image registration}
\author{Vincent Jaouen, Pierre-Henri Conze, Guillaume Dardenne, Julien Bert and Dimitris Visvikis, \IEEEmembership{Senior Member, IEEE}
\thanks{The authors are with LaTIM, INSERM, IMT Atlantique, CHRU Brest, Brest, France.
e-mail: vincent.jaouen@imt-atlantique.fr.}%
\thanks{This work benefited from state aid managed by the National Research Agency under the future investment program bearing the reference ANR-17-RHUS-0005.}}
\maketitle

\begin{abstract}

In image registration, many efforts have been devoted to the development of alternatives to the popular normalized mutual information criterion. Concurrently to these efforts, an increasing number of works have demonstrated that substantial gains in registration accuracy can also be achieved by aligning structural representations of images rather than images themselves. Following this research path, we propose a new method for mono- and multimodal image registration based on the alignment of regularized vector fields derived from structural information such as gradient vector flow fields, a technique we call \textit{vector field similarity}. Our approach can be combined in a straightforward fashion with any existing registration framework by substituting vector field similarity to intensity-based registration. In our experiments, we show that the proposed approach compares favourably with conventional image alignment on several public image datasets using a diversity of imaging modalities and anatomical locations.
\end{abstract}


\IEEEpeerreviewmaketitle

\section{Introduction}
Image registration (IR) is required in a large variety of medical imaging studies involving tasks such as inter-subject comparison \cite{Ardekani2005},  patient follow-up \cite{Brock2006}, modality fusion \cite{Heinrich2013}, atlas generation \cite{Lorenzen2006}, or more recently cross-modality image synthesis with deep learning techniques \cite{Florkow2019}. Automated IR is however a difficult problem, as medical images can be acquired from  modalities based on radically different physical principles and/or biological properties (e.g. anatomical imaging compared to functional imaging), using various image sampling strategies (e.g. slice thickness), showing inconsistent representation of underlying structures such as organs or tumours due to motion (e.g. respiratory movements \cite{Fayad2017}) or structural alterations (e.g. disease progression \cite{Zacharaki2009}). For these reasons, image registration is an active field of study followed by large communities of researchers and advanced users of dedicated software such as ANTs \cite{Avants2014} or \textit{elastix} \cite{Klein2010}. Over the recent years, the research in this field has mostly focused towards intensity-based ("voxel-based"), volumetric deformable image registration (DIR), a general alignment scenario accounting for nonlinear deformations between images and where no information other than voxel data is available \cite{Viergever2016}.

Many components in a DIR pipeline can have a strong influence on the accuracy of the results. The similarity metric (SM) is often regarded as the most critical \cite{Sotiras2013}, although a variety of other parameters such as the number of iterations of the optimizer or the optimizer itself \cite{Klein2009}, the number of scales in multiscale methods, or the nature of the regularizer \cite{Vishnevskiy2016} have also shown to strongly influence performance. 
Regarding the choice of the SM, normalized mutual information (NMI) \cite{Viola1997,Studholme1999} has remained highly popular for more than two decades \cite{Sotiras2013} due to its high success and its relative insensitivity to modality changes. Despite this NMI has also some disadvantages. For example, it is known to exhibit many local minima \cite{Haber2006} and to be sensitive to inhomogeneities such as bias fields in magnetic resonance images (MRI) \cite{Heinrich2012}. It is generally admitted that the limitations of NMI are mainly due to the global nature of the intensity histograms, as spatial relationships between voxels are not considered \cite{Loeckx2009,Heinrich2012,Viergever2016}. As pointed out by the authors of   \textit{elastix} "\textit{it is unlikely that mutual information will be able to maintain its popularity, given the need for local measures of image similarity}" \cite{Viergever2016}. A number of studies have therefore been devoted to the development of more robust, local and modality-independent alternatives to the NMI criterion \cite{Haber2006,Loeckx2009,Heinrich2012,Simonovsky2016}. 

Among these efforts towards the development of new SM, Haber and Modersitzki proposed the normalized gradient field (NGF) \cite{Haber2006}. The rationale of NGF is to maximize the alignment between pairs of normalized image gradient vectors (instead of intensity images). The sign of the vectors are ignored to enforce robustness to contrast inversion between different modalities. The normalization step ensures that contrast variations across regions do not influence registration results. Similar concepts were introduced  earlier in \cite{Pluim2000} where gradient similarity was used in combination with mutual information-based intensity registration. In practice, NGF is implemented as a SM through a variational formulation by either maximizing the square of the dot product between the two fields, or equivalently by minimizing the norm of the cross product.  A limitation is that normalized gradients are only meaningful around edges, whereas in homogeneous regions their directions depend on local noise. To alleviate this issue, an empirical gradient threshold is defined based on the estimated noise level to suppress the influence of weak gradients attributable to noise. However, thresholding may not be optimal as it discards weak but real edges and preserves high amplitude noise. Another recently popular SM is the modality independent neighborhood descriptor (MIND) \cite{Heinrich2012}, which encourages local structural self-similarity. The local structure is in itself described by a vector-valued representation of the local similarity of small image patches. Such descriptors can then be compared between two images using conventional monomodal SM like the sum of square differences (SSD). However, the MIND does not encode the spatial orientations of image structures, contrary to NGF. Both NGF and MIND are being increasingly used in the community. For instance, NGF is used in the current winning method of the EMPIRE10 lung computed tomography (CT) DIR challenge \cite{Murphy2011}, followed by a method based on MIND. A more recent research path is to learn DIR unsupervisedly from training data using deep neural networks \cite{Simonovsky2016}. However, deep learning-based strategies for IR are still in their infancy \cite{Blendowski2020}, and it is not clear from recent results that they can outperform more conventional optimization approaches based on handcrafted similarity metrics and user software experience \cite{Devos2019}, as it is the case for example in supervised image segmentation with deep strategies such as U-Nets \cite{Ronneberger2015}.

Another way of improvement concurrent to the development of new SM focuses on \textit{structural representations} (SR) \cite{Wachinger2012}, whereby images themselves are transformed before being aligned in such a way that conventional monomodal SM, like the SSD, can be employed. The authors of \cite{Wachinger2012} introduce \textit{Entropy} and \textit{Laplacian} images,  two SR that offer theoretical guarantees in terms of {locality preservation} and {structural equivalence} at the patch level. They demonstrate these SR can achieve superior performance over the use of intensity images. The motivations for modality independent SR are similar to the ones developed for the MIND approach, and identical concepts are now being adopted in deep learning-based DIR under the name of \textit{image transformer networks} \cite{Lee2019,Arar2020}, whereby a geometry preserving image translation neural network is trained to learn a SR, which is then fed to a \textit{spatial transformer network} \cite{Jaderberg2015} in charge of the actual image warping.  More generally and relaxing theoretical requirements at the patch level, SR can be seen as a form of domain adaptation ensuring that heterogeneous data can be compared while suppressing modality-specific image characteristics. As pointed out by the authors of MIND \cite{Heinrich2012}, the SR in \cite{Wachinger2012} are scalar-valued. As such, they cannot convey directional information relative to the orientation of image structures. The MIND descriptor captures local self-similarity in a vector-valued SR, but it also does not carry information related to structure orientations. This property is an essential advantage of the NGF metric over these SR-based approaches.

In this work, we propose a new method for multimodal image registration based on the evaluation of the similarity between regularized directional \textit{vector-valued} fields derived from structural information, a technique we call Vector Field Similarity (VFS). These new representations are derived from edge-based fields (EBF) that are usually used for deformable model segmentation purposes to provide smooth vector fields oriented towards edges \cite{Xu1998,Li2007,Jaouen2014}. We show that EBF, by encoding regularized edge orientation in a vector-valued SR, can overcome the limitations of intensity-based registration. Similarly to existing SR-based registration methods, our approach can be combined in a straightforward and flexible fashion with any vector-valued registration framework regardless the SM chosen. As already noted in \cite{Heinrich2012}, this is in contrast to the NGF approach, which is formulated at the metric level. Amongst other advantages, SR enable a direct and fair comparison with existing registration pipelines, by using SR as substitutes for intensity images, all else being equal. In our experiments, we show that VFS compares favorably to conventional intensity-based methods on public datasets, using multiple image registration scenarios for a variety of imaging modalities and anatomical locations. 

\section{Methods}
\begin{figure}[htbp]
\begin{center}
\subfigure[CT image]{
\includegraphics[width=.45\linewidth]{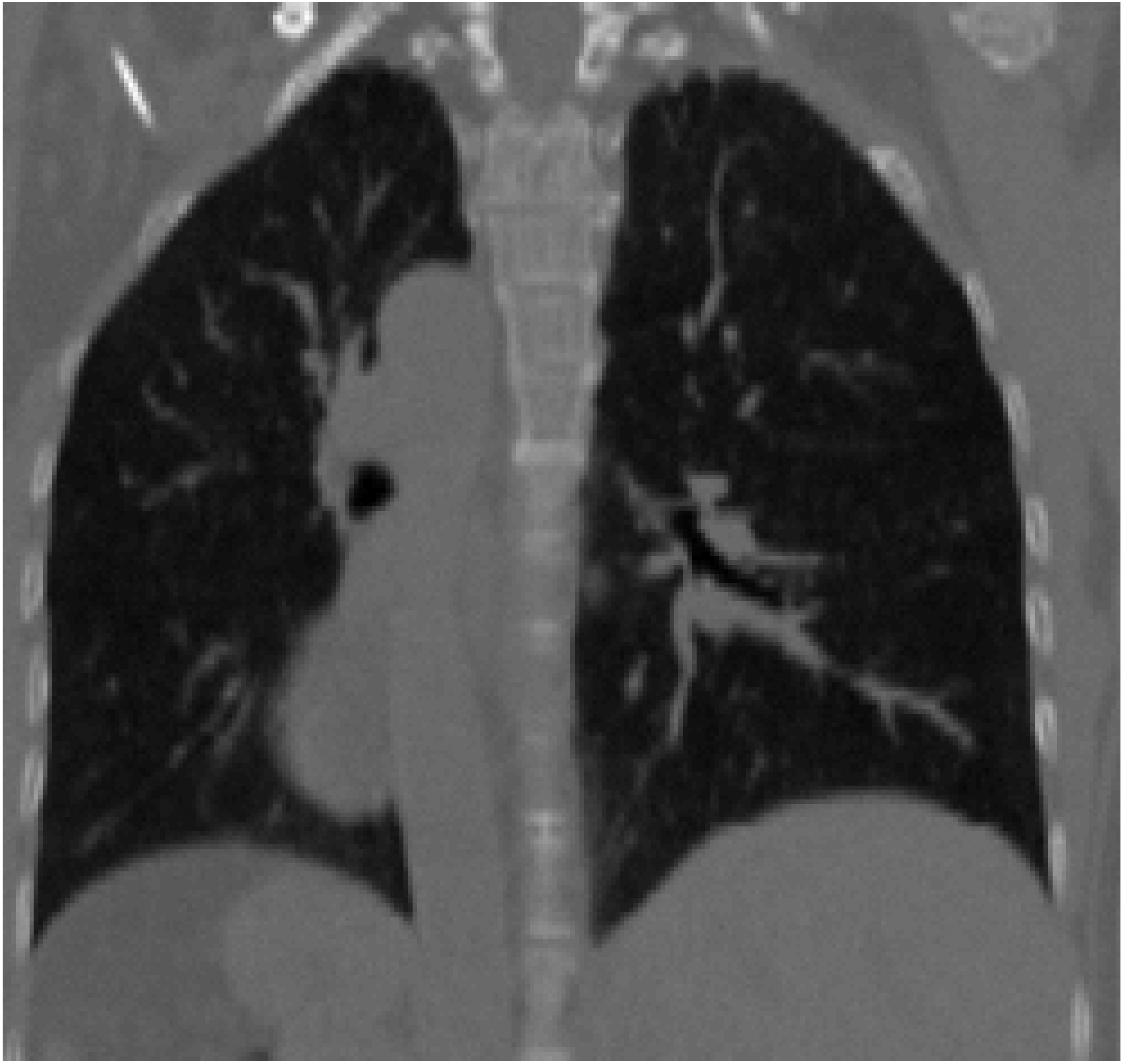}
}
\subfigure[NGF]{
\includegraphics[width=.45\linewidth]{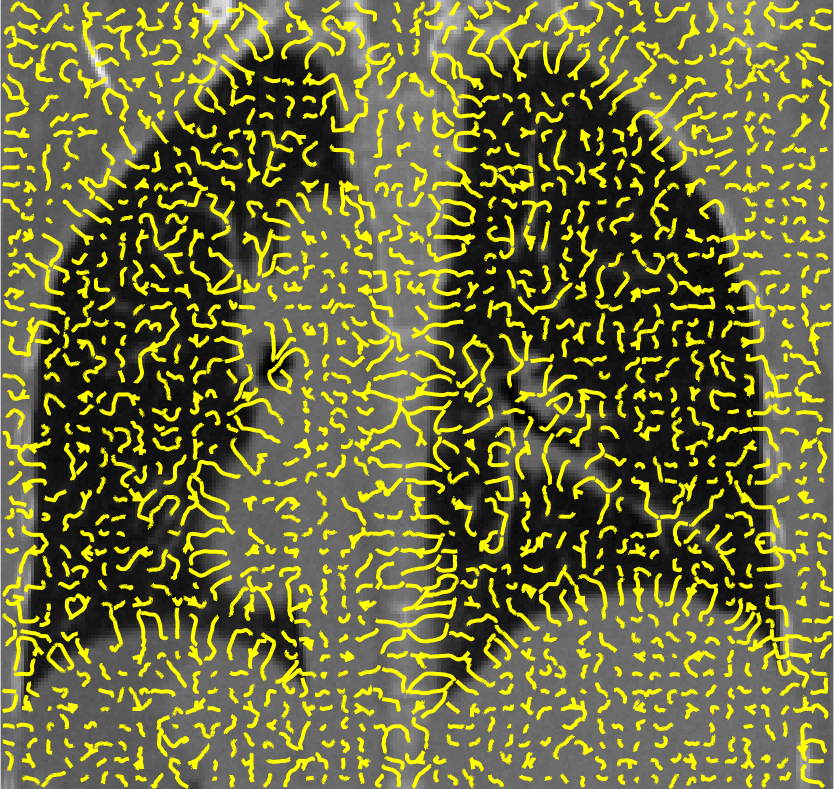}
}
\subfigure[EBF, $\gamma=5.0$]{
\includegraphics[width=.45\linewidth]{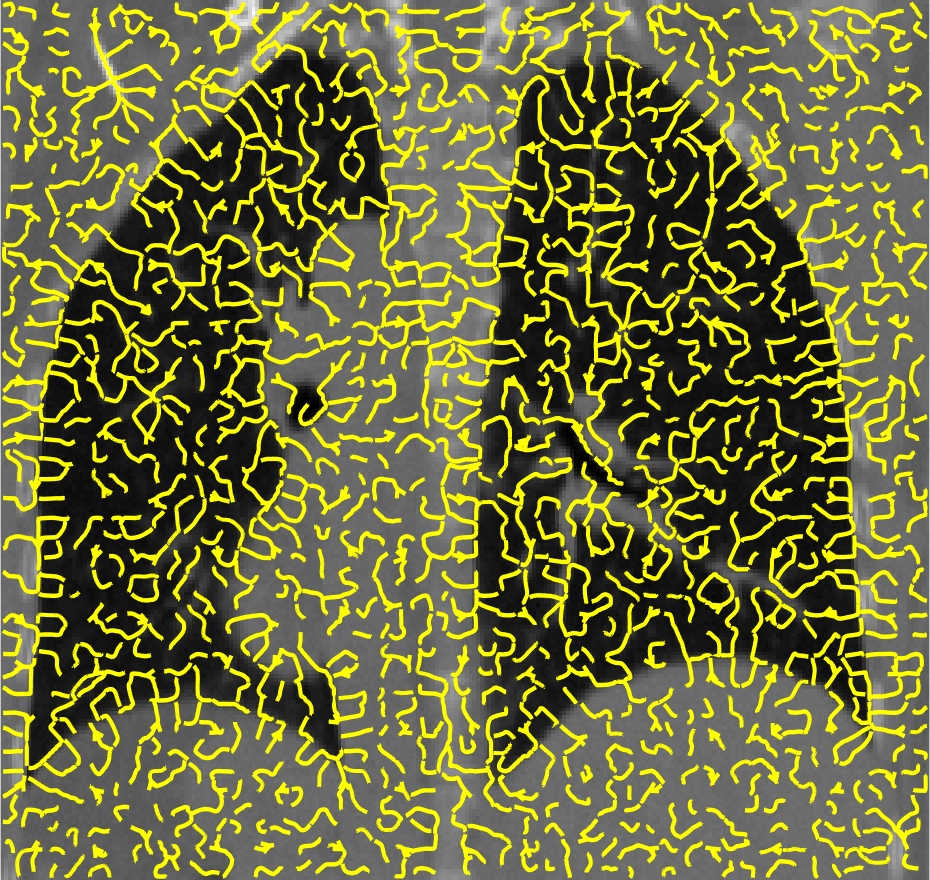}
}
\subfigure[EBF, $\gamma=4.0$]{
\includegraphics[width=.45\linewidth]{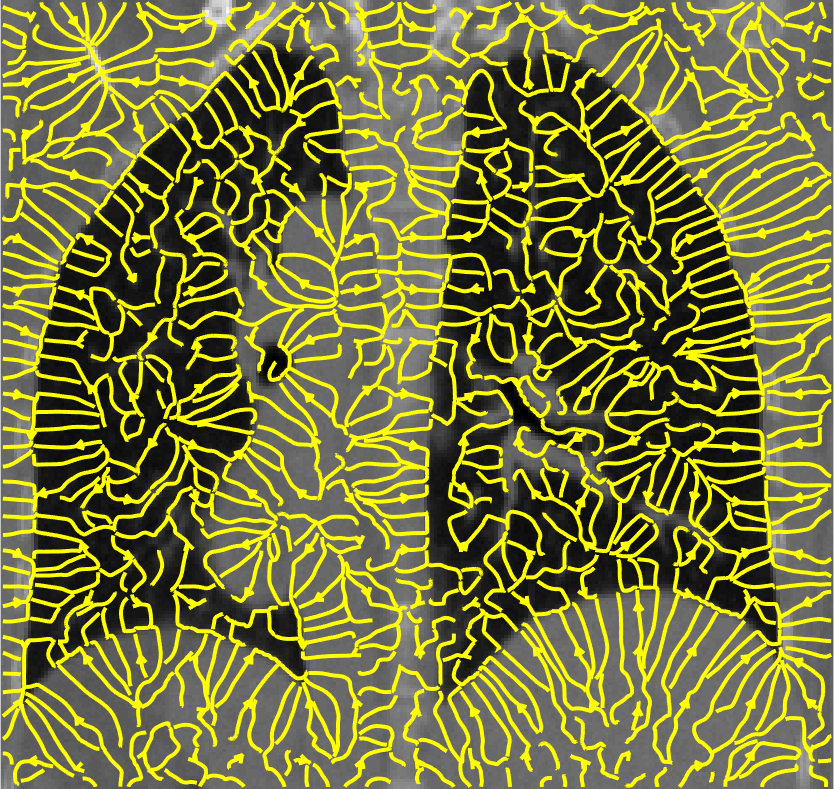}
}
\subfigure[EBF, $\gamma=3.0$]{
\includegraphics[width=.45\linewidth]{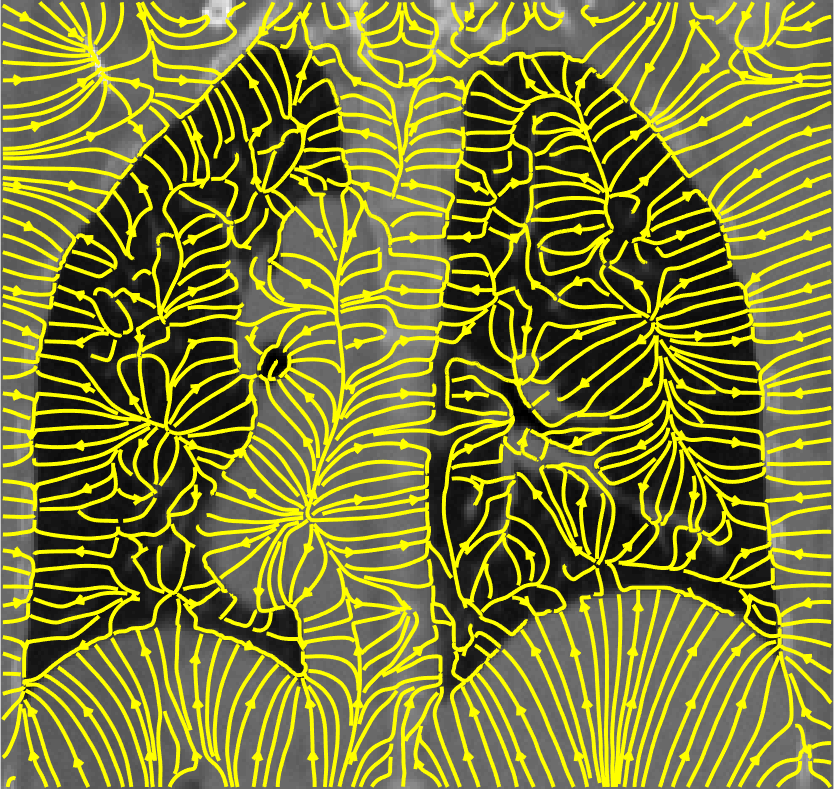}
}
\subfigure[EBF, $\gamma=2.0$]{
\includegraphics[width=.45\linewidth]{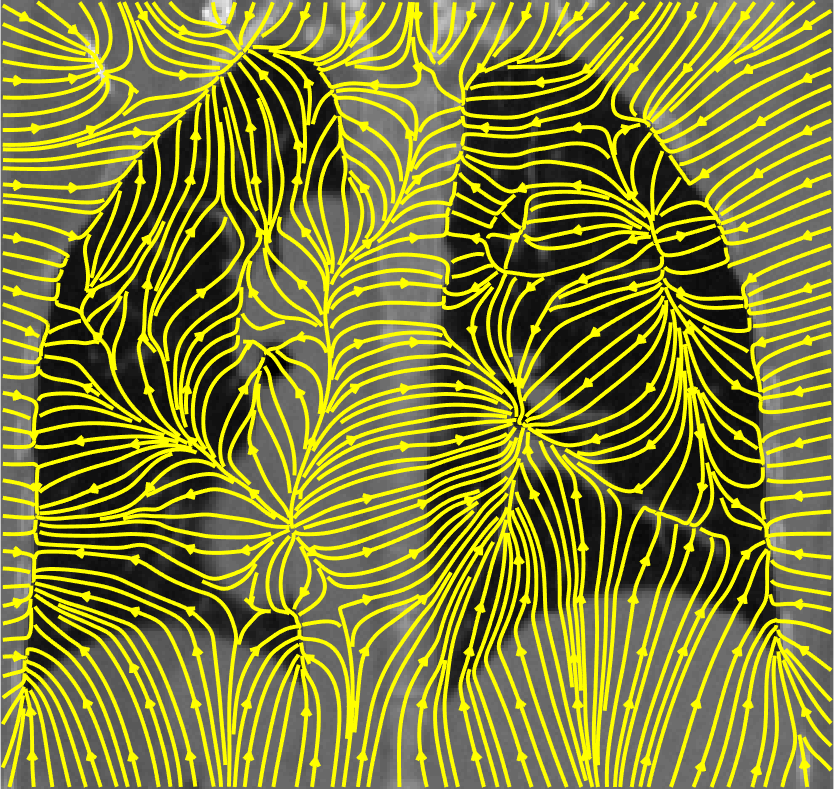}
}
\end{center}
\caption{\label{figVFC} Differences between normalized gradient fields (NGF) and edge-based fields (EBF) shown in a patient of the DIR-Lab dataset. The EBF is a vector field convolution field \cite{Li2007}. Results are shown using streamlines to ease visualization. (b) NGF field (c-f) EBF field with increasing levels of smoothing, smaller $\gamma$ indicates stronger smoothing. For high values of $\gamma$, NGF and EBF are similar. For stronger smoothing effect, EBF vectors are directed towards edges including in homogeneous regions,  contrary to NGF. Near edges, EBF orientation is independent of contrast inversion and points towards edges.}
\end{figure}

\subsection{Vector Field Similarity}

We consider a fixed image $I$ and a moving image $J$ defined on the image grid $\Omega$. The aim of the proposed image registration method is to find a transformation $\hat T$ such that: 
\begin{equation}
\hat T = \argmax_{T \in \mathcal{T}} ~ \mathcal S\left(\mathbf D^I,\mathbf D^J\left(T\right)\right), 
\end{equation}
where $\mathcal S$ is a similarity metric such as NMI or SSD, $\mathcal{T}$ the space of transformations, and $\mathbf D^I$ and $\mathbf D^J$ are {vector-valued} SR of $I$ and $J$. This is different from the method proposed in \cite{Wachinger2012} where {scalar} SR are considered. 

Several strategies can be considered to define $\mathbf D$, one of which is the MIND descriptor \cite{Heinrich2012}, which encodes local self-similarities. The NGF method is on the other hand expressed at the metric level $\mathcal S$ and describes the similarity between the normalized gradient fields of the fixed and the moving images. One of its formulation  is:
\begin{equation}
\mathcal S_\text{NGF} := -\frac{1}{2}\int_{\mathbf x \in \Omega}  \langle \mathbf n_\varepsilon\left(I\right), \mathbf n_{\varepsilon}\left(J\left(T\right)\right)\rangle ~\dd{} \mathbf x,
\end{equation}
where $\mathbf n_\varepsilon(I)$  and $\mathbf n_\varepsilon(J)$ are normalized gradient fields of the fixed and moving images: \cite{Haber2006}:
\begin{equation}
\mathbf n_\varepsilon\left(I\right) :=\frac{\nabla I}{\sqrt{\nabla I^T \nabla I+\varepsilon^2}},
\end{equation}
with $\varepsilon$ an estimate of the noise level of $I$.

 A main advantage of NGF over local descriptors such as MIND is that it encodes the orientations of image structures identified by the image gradient. In principle, normalized gradient vectors can be directly considered as a vector-valued structural representation rather than at the metric level.
Here, we propose to use edge-based vector fields (EBF) normally used for image segmentation for this purpose, a technique we call Vector Field Similarity (VFS).

 EBF such as the popular gradient vector flow \cite{Xu1998} are smooth vector fields derived from edge information and oriented towards edges \cite{Xu1998}. EBF enable to extend the capture range of deformable models and to reduce sensitivity to noise through a regularization of the field orientations in homogeneous, edge-free regions (Fig. \ref{figVFC}c-f). They can also show some contour completion abilities in case of missing edges \cite{Jaouen2019,Jaouen2019a}. These properties are not only desirable for image segmentation with active contours, but also for image registration purposes and may help overcoming some of the limitations of previously proposed gradient-based alignment methods, such as  NGF. For example, EBF orientations in homogeneous regions contribute favorably to image alignment, contrary to NGF where vectors point in random directions due to noise (Fig. \ref{figVFC}b). 
 
\subsection{Vector Field Convolution fields}

\begin{figure}[h]
\centering
\subfigure[T1-weighted BrainWeb image.]{
\includegraphics[width=.45\linewidth]{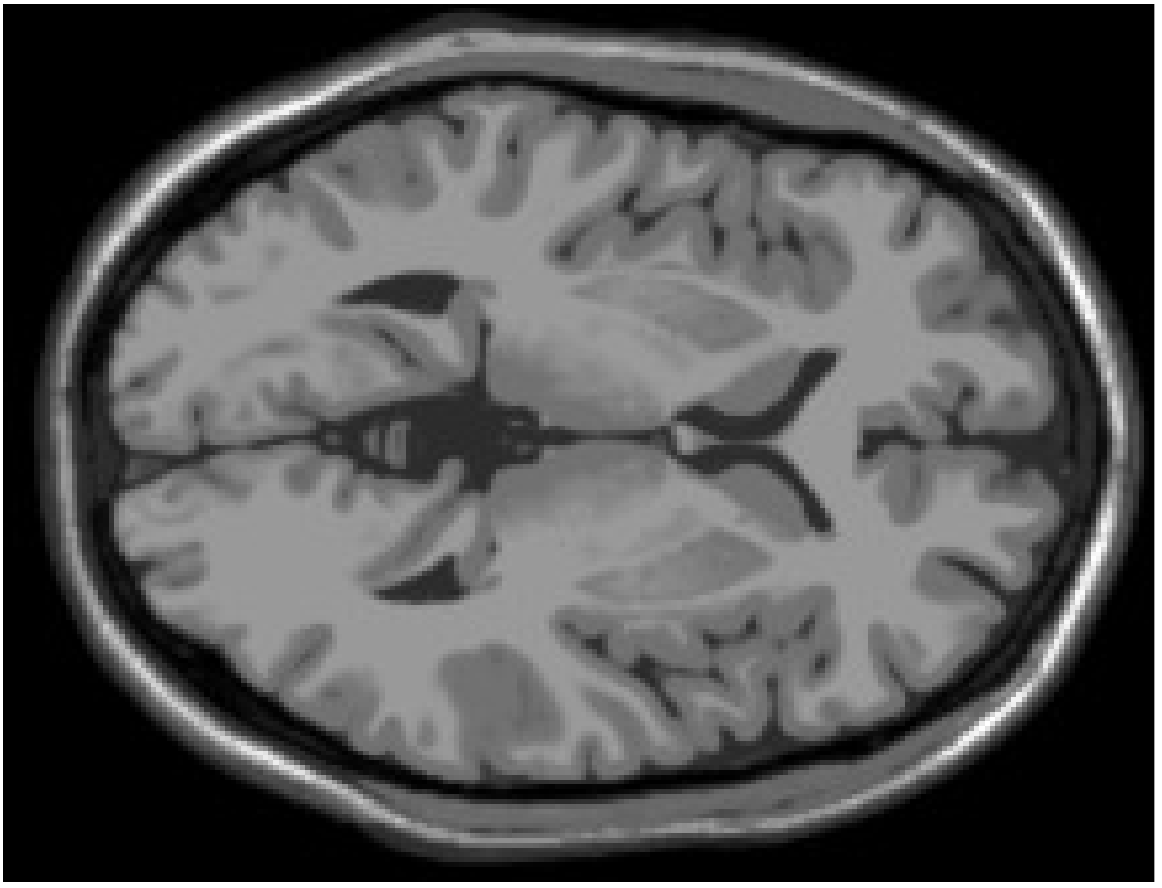}
}
\subfigure[T2-weighted BrainWeb image.]{
\includegraphics[width=.45\linewidth]{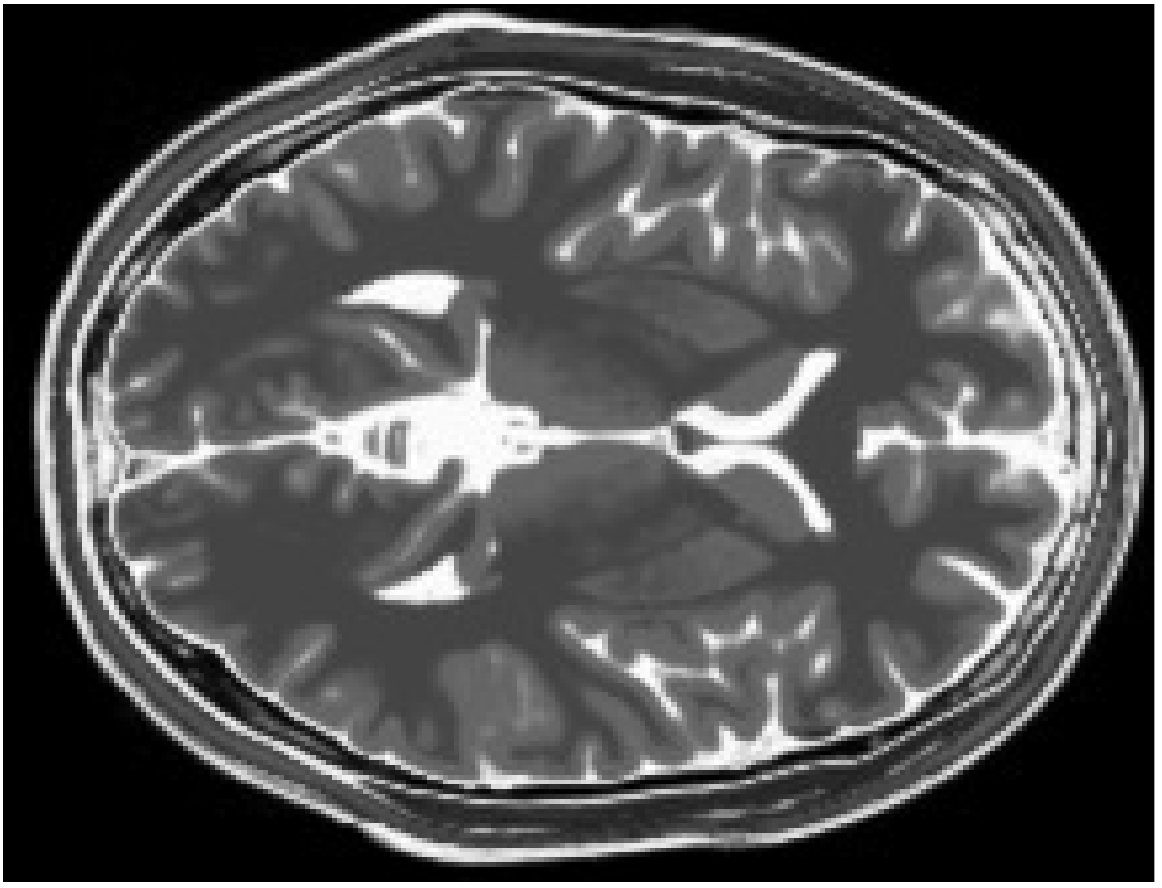}
}
\subfigure[Noise level: $0\%$]{
\includegraphics[width=.9\linewidth]{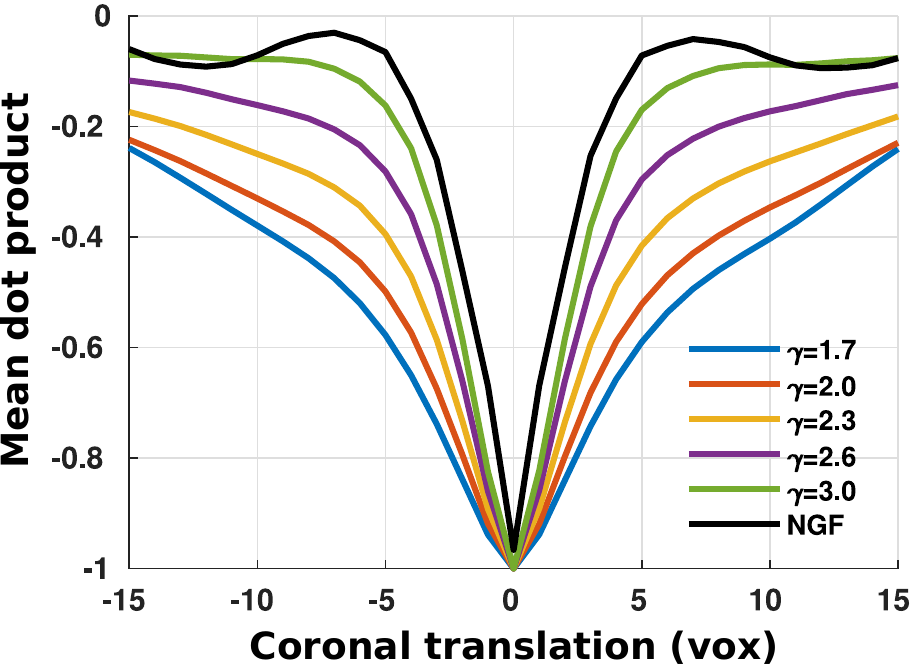}
}
\subfigure[Noise level: $9\%$]{
\includegraphics[width=.9\linewidth]{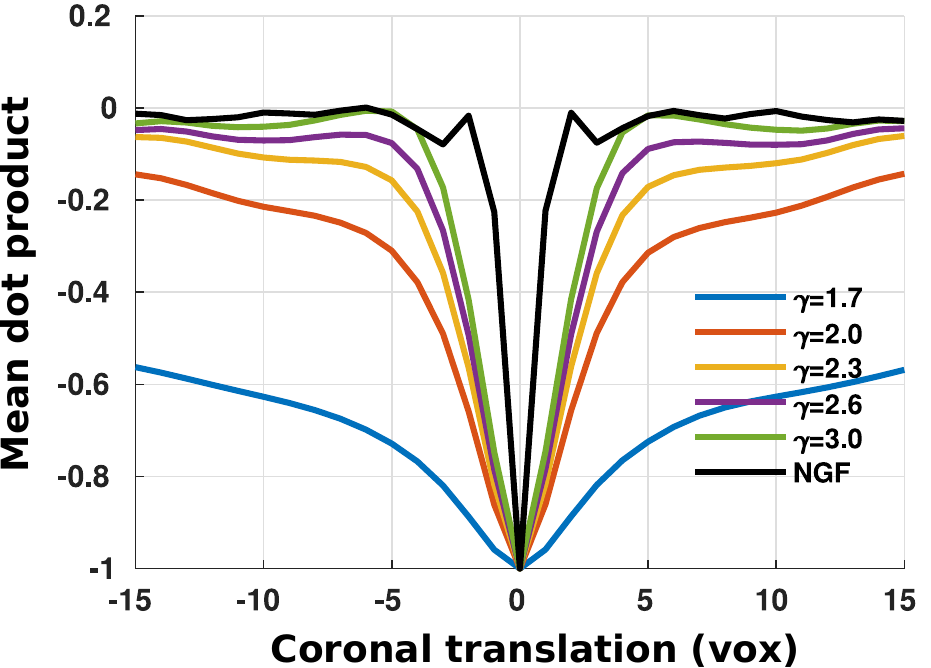}
}
\caption{\label{figTranslationMono} Effect of noise on a self translation study along the coronal axis using the T1-weighted BrainWeb image shown in (a) for both normalized gradient (NGF) and vector field convolution similarity (VFS). The value is the average norm of the dot product, negated by convention to show attraction basins. VFS results are shown for several values of the nonlinear smoothing VFC parameter $\gamma$.}
\end{figure}

\begin{figure}[h]
\centering
\subfigure[Noise level: $0\%$]{
\includegraphics[width=.90\linewidth]{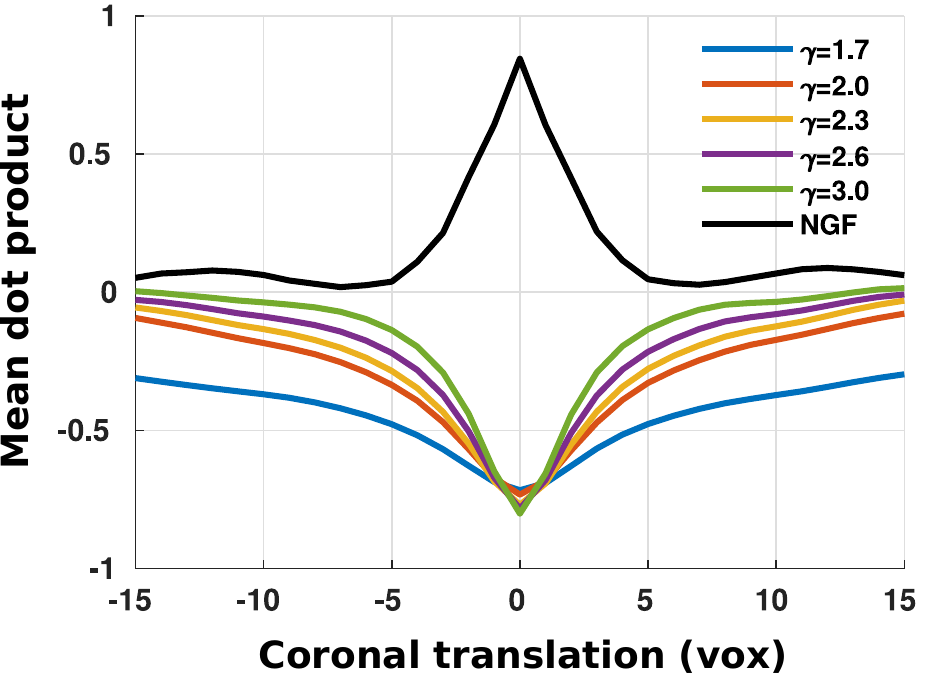}
}
\subfigure[Noise level: $9\%$]{
\includegraphics[width=.90\linewidth]{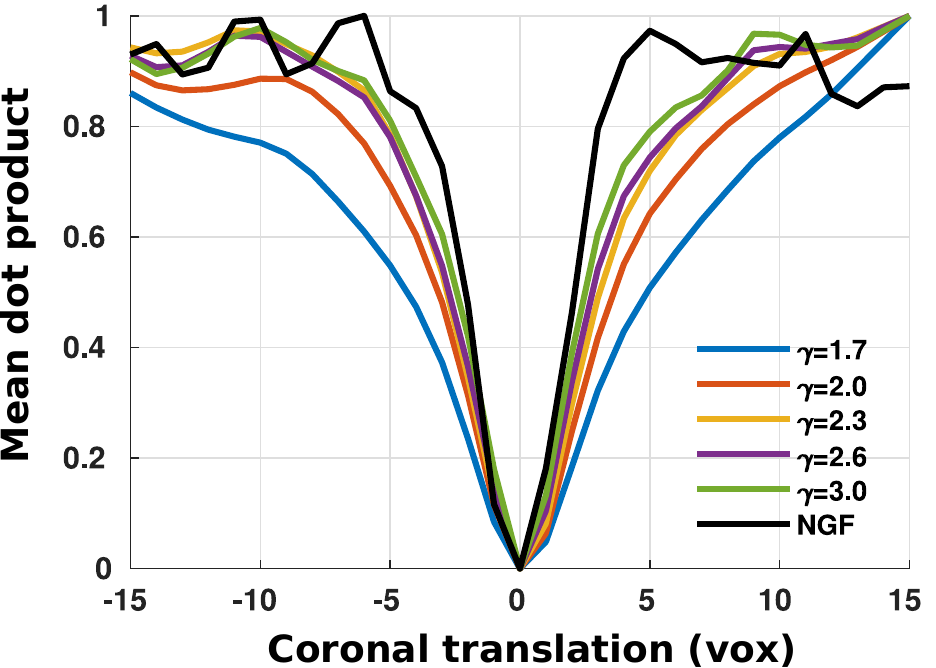}
}
\caption{\label{figTranslationMulti} Multimodal (multiparametric) translation study between the T1 and T2 images of the BrainWeb database. Due to contrast inversions between the modalities the NGF sign in (a) is reverted compared to Fig \ref{figTranslationMono}c. To ease and provide further elements of visual comparison, we negate the values for NGF in (b) scale all profiles between 0 and 1 .}
\end{figure}

In the present work, without loss of generality, we choose vector field convolution (VFC) \cite{Li2007} to generate robust and smooth EBF. VFC fields share several desirable properties with GVF fields such as large capture range while being more computationally efficient, requiring only $n$ convolutions with a kernel in $n$-dimensional images \cite{Xu2020}. This property is especially useful when considering large volumetric medical images. VFC fields also demonstrate superior robustness to impulse noise compared to GVF \cite{Li2007}.  A VFC field $\mathcal F$ is expressed as:  
\begin{equation}
\mathcal F = f * \mathcal K,
\end{equation}
where $f$ is an edge map typically derived from structural information like image gradient and $*$ is the convolution operation. In our experiments, we chose ${f=\left\|\nabla I\right\|^2}$ for simplicity, but more elaborate approaches such as edge maps based on local structure tensors \cite{Jaouen2014} or  image pre-processing \cite{Bazan2007} can naturally be considered. $\mathcal K$ is a \textit{vector field kernel}, a vector kernel whose orientations point towards its center with decreasing magnitude $m$: \begin{equation}
m = \frac{1}{r^\gamma+\epsilon}
\end{equation} where $r$ is the distance to center \cite{Li2007}, $\epsilon$ a small positive value to avoid division by zero at the center, and $\gamma$ a fixed parameter that controls the rate of decrease. 

 To illustrate the properties of VFC fields, Fig. \ref{figTranslationMono} shows a self translation study using the BrainWeb \cite{Cocosco1997} T1-weighted MRI (Fig. \ref{figTranslationMono}a) for two distinct noise levels ($0\%$ and $9\%$, corresponding to the minimum and maximum noise levels of the BrainWeb database). Results with increasing values for the smoothing parameter $\gamma$ are shown for VFS. The degree of co-alignment between the image and its translated version is measured as the average dot product between the fixed and the moving fields using either normalized VFC fields or NGF fields, i.e. perfect alignment is of magnitude $1$. To respect registration conventions, the negative of the similarity value is shown. Even in the idealized case of self-registration without noise (Fig. \ref{figTranslationMono}c), the attraction basin of the NGF becomes quickly non convex. This is due to the inversion of the gradient signs even at small transation shifts. This non-convexity increases with noise, narrowing the capture range of gradient similarity to a few voxels (Fig. \ref{figTranslationMono}d). The dependency on the sign of the vectors of NGF can be lifted by squaring the magnitude of the dot product, as proposed in previous related works \cite{Haber2006,Pluim2000}. However, squaring first-order gradient information leads to noise amplification. On the contrary,  VFS alleviates this need while allowing tunable nonlinear smoothing to further reduce sensitivity to noise. VFS also more essentially provides some form of feature selection by accumulating  the weight of relevant edges while smearing out noise \cite{Li2007}. The effect of VFS on image registration is mainly determined by the degree of smoothness of the EBF field. In the case of VFC, this smoothness is controlled by a unique parameter $\gamma$, which specifies the tradeoff beween weak edge preservation and noise reduction. The higher $\gamma$ the lesser smoothing of the field occurs, and VFC field similarity becomes similar to NGF field similarity. More elaborate EBF can also be considered depending on the context, such as generalized GVF fields \cite{Xu1998a} or fields dedicated to vector-valued images \cite{Jaouen2013,Jaouen2014} that may use additional parameters to better tune edge preservation (through e.g. gradient thresholding or nonlinear structure tensor smoothing). However in the present case of VFC-based VFS registration $\gamma$, is a unique  additional parameter that needs be tuned to a specific registration application that act as a tradeoff between robustness to noise and preservation of local orientations of image structures.
 
 Fig. \ref{figTranslationMulti} shows a  translation study similar to Fig. \ref{figTranslationMono} in a multimodal context, where the translated image is replaced by the T2-weighted BrainWeb image (Fig. \ref{figTranslationMono}b). Due to global contrast inversion between T1 and T2 weightings, the sign of the NGF is flipped with respect to Fig. \ref{figTranslationMono} (Fig. \ref{figTranslationMulti}a). Also, the maximum alignment value no longer achieves a magnitude of $1$ as the images are not identical. The lack of convexity of the attraction basin of the NGF becomes more prominent with added noise (Fig. \ref{figTranslationMulti}b), while on the contrary VFS consistently provides a smoother energy landscape for various values of $\gamma$. 
 
Contrary to the original NGF approach, which is formulated at the similarity metric level and where the alignment of gradient vectors are directly optimized based on dedicated alignment metrics, we consider EBF to be vector-valued structural representations of the underlying images, similarly to the philosophy behind (scalar-valued) entropy or Laplacian structural representations \cite{Wachinger2012} or the (vector-valued) MIND approach \cite{Heinrich2012}. This point of view enables the use of any other SM deemed relevant to the problem at hand to evaluate VFS, such as e.g. SSD, NMI or normalized cross correlation (NCC).

\section{Evaluation process and datasets}

The difficulty of validating image registration methods has been largely discussed and argumented by experts in the field in the last decade \cite{Murphy2011, Rohlfing2013,Viergever2016}. A consensus is that the preferred validation strategy should be to use dedicated datasets carefully annotated with dense point landmarks (of the order of several hundreds per organ), and to rely on average distance-based metrics such as mean target registration errors (TRE). This type of ground truth is naturally extremely difficult to produce and is, as of today, only available publicly for lung CT studies \cite{Castillo2009,Castillo2013,Murphy2011}. Due to lack of annotations, using surrogates to landmark-based validation is nevertheless often the only validation strategy available and should be done carefully. A study by Rohlfing showed that a great number of DIR studies published in top journals and international conferences did not respect validation standards, relying too heavily on validation surrogates such as unsupervised image similarity metrics or volumetric overlaps between insufficiently small anatomical structures \cite{Rohlfing2013}.  These observations are still valid for many more recent studies.  

Another challenge with DIR validation is to demonstrate the  superiority with respect to the state-of-the art. DIR pipelines contain a large quantity of components that can have a critical influence on the results \cite{Klein2010,Modersitzki2009,Tustison2014}, and success is most often achieved through intensive practice  and extensive parameter exploration. Due to the size of the parameter space and to nonlinear interdependencies between these parameters, it is however extremely difficult to establish a fair ranking among different methods \cite{Murphy2011}. This task is nevertheless facilitated by active communities of software users such as \textit{elastix} \cite{Klein2010} or ANTs \cite{Tustison2014} and challenge organizers sharing best practice and pitfalls associated with various use cases for different anatomical locations, the most common of which being the brain and the lungs.

To partly address these issues raised with the validation of image registration algorithms, we evaluated the proposed VFS-based registration approach using existing DIR pipelines that were previously validated by research groups on the same public image datasets that we used in our experiments when available. All else being equal, we evaluated the change in performance using VFS by substituting EBF-based SR to the intensity image. That is, instead of the image-based similarity metric $\mathcal S  \left(I, J\right)$ used in a given pipeline, we averaged the same metric on the $n$ components of the EBF:
\begin{equation}
\mathbf S \left(\mathbf D^I, \mathbf D^J\right) = \frac{1}{n} \sum_{i=1}^n \mathcal S \left(D_i^I, D_i^J\right),
\end{equation}
where $D_i^I$ and $D_i^J$ are the i$^\text{th}$ components of $\mathbf D^I$ and $\mathbf  D^J$ ($n=3$ in volumetric images).  

In this context, rather than absolute performance, the objective is to demonstrate that regardless the pipeline considered, substituting vector structural representation to the intensity image leads to improvedresults in several registration scenarios. In addition, this constrained validation setup can ease comparative evaluation by removing potential bias due to algorithm re-implementation. Such a choice is not optimal for VFS that could likely benefit from further optimization. 

In all our experiments, $\mathbf D^{I,J}$ are VFC fields computed using a discrete kernel support size of $100$ voxels. For practical reasons, we implemented our own version of VFC in C++ using ITK \cite{Avants2014}. A Matlab implementation provided by the authors of VFC is available online\footnote{http://viva-lab.ece.virginia.edu/pages/toolbelt.html}.  For all datasets, the VFC exponent parameter $\gamma$ controlling the degree of smoothness of the EBF was empirically selected in the range $[2.5,4.5]$ so as to maximize the studied evaluation metric and is specified for each experiment in the next section. 

We used several publicly available datasets covering diverse anatomical locations including brain MRI and lung CT imaging, which are two common applications for image registration. We also propose the use of a publicly available abdominal MRI segmentation dataset which is also well-suited as a surrogate for the validation of multimodal image rgistration since it focuses on small anatomical structures (kidneys). The rationale behind the choice of the datasets was to cover a large variety of registration scenarios while encouraging reproducibility. More specifically, we used both Hammers\footnote{https://brain-development.org/brain-atlases/adult-brain-atlases/} \cite{Hammers2003} and IBSR18\footnote{https://www.nitrc.org/projects/ibsr} brain MRI datasets, the DIR-Lab\footnote{https://www.DIR-Lab.com/} thoracic CT datasets \cite{Castillo2009} as well as the  CHAOS\footnote{https://chaos.grand-challenge.org/} abdominal multiparametric MRI segmentation dataset \cite{Chaos2020}, which we propose to use for the first time for image registration purposes.

\subsection{\label{sec3a} Landmark-based validation on lung 4DCT images} 

The only public datasets providing dense landmark correspondences are the three DIR-Lab studies  \cite{Castillo2009,Castillo2009a,Castillo2013}, and the EMPIRE10 challenge dataset \cite{Murphy2011}, which all concern lung 4DCT  studies. The validation of EMPIRE10 results is indirect as it requires a written submission and a participation to the challenge website. In our experiments, we focused on the three DIR-Lab datasets, which provide access to $300$ landmarks' correspondences between the end-inhale and end-exhale phases for $20$ patient cases. The DIR-Lab datasets are divided in $3$ sub-datasets, hereafter referred to as studies DIR-Lab-4DCT-1 \cite{Castillo2009}, DIR-Lab-4DCT-2 \cite{Castillo2009a} and DIR-Lab-COPd \cite{Castillo2013} showing variations in displacement amplitudes between the two breathing phases. 

We evaluated VFS in the context of thoracic imaging using two existing dedicated registration pipelines: 
\begin{itemize} 
\item a conventional \textit{elastix} DIR pipeline, hereafter referred to as ELX, constituted of three consecutive stages (one affine and two B-spline stages using the NCC metric) that was primarily designed for the EMPIRE10 lung registration challenge where it achieved the second best ranking at the time of submission \cite{Murphy2011}. An \textit{elastix} parameter file\footnote{http://elastix.bigr.nl/wiki/index.php/Par0011} was released by the authors allowing for an objective comparison with VFS. The second DIR stage (ELX-2) is used to refine results from the first DIR stage (ELX-1) using lung masking. In the following, the corresponding VFS-based results are referred to as VFS-1 and VFS-2 respectively.

\item the \textit{pTV} \footnote{https://github.com/visva89/pTVreg} pipeline based on isotropic total variation regularization \cite{Vishnevskiy2016}, which is currently the  best performing method on the DIR-Lab datasets. The rationale behind pTV is to allow for sharp transitions in the  deformation fields along sliding interfaces through total variation-based regularization of the control grid displacements of first order B-splines. Such transitions are common in the lung and impossible to render with $\ell_2$-based smoothness penalization. We found the mean TRE values reported in \cite{Vishnevskiy2016} to be slightly sub-optimal when running the algorithm and we therefore report two sets of measurements, referred to as pTV-1 for values reported in \cite{Vishnevskiy2016}  and pTV-2 for the updated values.

\end{itemize}
In this experiment, the VFS parameter $\gamma$ was set to $\gamma=3$.
\subsection{\label{sec3b} Subcortical volume overlaps in MRI}

A major recommendation of Rohlfing's study \cite{Rohlfing2013}  is that, in the absence of landmark-based correspondences, "\textit{only overlap of sufficiently
local labeled ROIs could distinguish reasonable from poor registrations}", the author arguing that \textit{"smaller, more localized ROIs
approximate point landmarks, and their overlap thus approximates point-based registration error"}. These principles have guided our validation strategy in the remainder of this section.

The Hammers dataset \cite{Hammers2003} is a brain T1-weighted MRI dataset consisting of $30$ subjects that were manually segmented into $83$ regions. Similarly to section \ref{sec3a}, we compared our approach to an existing study by Qiao et al. \cite{Qiao2015} for which parameter files for are also available on the \textit{elastix} online database website \footnote{http://elastix.bigr.nl/wiki/index.php/Par0035}. We reproduced the results described in \cite{Qiao2015} by performing inter-subject registration for all patients, leading to 870 fixed-moving image pairs. For each subject, we studied the average Dice similarity coefficient (DSC) in the sixteen labeled subcortical regions (left and right hippocampus, amygdala, cerebellum, caudate nucleus, nucleus accumbens, putamen, thalamus and pallidum) after image registration with the $29$ remaining images. We focused on subcortical structures in agreement with the recommendations of Rohlfing \cite{Rohlfing2013}. Although the cerebellum arguably does not meet these recommendations due to its large volume, it was kept in our experiments for completeness. Left and right segmentations were merged into bilateral segmentations for clarity, leading to $8$ labeled regions. A similar study was also conducted on the IBSR18 brain dataset consisting of $18$ healthy brain MRI subjects. The VFS parameter $\gamma$ was set to $\gamma=4$.

\subsection{\label{sec3c}Renal volume overlaps in multiparametric abdominal MRI}

We are also interested in the validation of VFS in a multimodal context. As of today, the only public multimodal dataset dedicated to image registration is the RIRE database for rigid brain alignment\footnote{http://www.insight-journal.org/rire}. To diversify anatomical locations, we propose the use of the recent Combined Healthy Abdominal Organ Segmentation (CHAOS) database \cite{Chaos2020}, a multiparametric MRI dataset for abdominal organ segmentation. Abdominal MRI is a less studied and nevertheless challenging problem for image registration \cite{Carrillo2000}. The CHAOS database is constituted of a multiparametric MRI study showing the segmentation of four abdominal organs (liver, spleen, right and left kidneys) acquired with two different pulse sequences (T1-DUAL and T2-SPIR) for $20$ patients that were neither co-registered nor resliced into the same image sampling space. In addition to showing a less explored anatomical location, several characteristics of the CHAOS dataset suggest its potential interest as a surrogate for image registration validation. A major advantage is that the manual ground truth segmentation masks of the different organs were performed independently in the two modalities. The kidneys are also relatively small organs compared to the field of view (FoV), which follows the recommendations of \cite{Rohlfing2013}. Finally, the relative symmetry of the kidneys enable a bilateral evaluation of the registration performance across the FoV. 

For these reasons, we have studied the average Dice overlap measurements for the left and right kidneys in the CHAOS dataset as a new surrogate measurement for multimodal image registration accuracy. As previously, we used an existing registration pipeline based on \textit{elastix}\footnote{http://elastix.bigr.nl/wiki/index.php/Par0057} (referred to as ELX)  for multiparametric MRI to evaluate VFS objectively. This pipeline was originally proposed in \cite{Jansen2019} for the registration of diffusion weighted MRI to abdominal dynamic contrast enhanced MRI for liver segmentation and tumor detection. It is a two-stage registration procedure (rigid then B-spline) and uses NMI as a similarity metric. The VFS parameter $\gamma$ was set to $\gamma=2.5$.

\section{Results}

\subsection{\label{sec4a} Landmark-based validation on lung 4DCT images}

Table \ref{tab1} and \ref{tab2} show results for the ELX comparative study on the three DIR-Lab datasets. Only the results for the final B-spline stage are shown in table \ref{tab2} for clarity. A consistent improvement of TRE was observed for all patients with the simple substitution of intensity images  for directional structural VFC-based representations. After the first stage, an average mean TRE of $3.34\pm1.77$ mm was achieved for ELX-1 against $2.23\pm1.06$ mm for VFS-1. Results were further improved with the second B-spline stages, with an average mean TRE of $2.17\pm1.04$ mm for ELX-2 (performing only slightly better than the first VFS stage) against $1.84\pm0.64$ mm after the second VFS stage.
Table \ref{tab3} shows mean TREs for the DIRLAB-4DCT datasets using isotropic total variation regularization. Results were more contrasted based on the dataset considered, with significant improvement for VFS in $4$ out of $5$ cases of study \cite{Castillo2009}. The substitution for VFS sets two new record TRE scores in the DIR-Lab challenge leaderboard for the first two cases (mean TREs of $0.69$ mm and $0.67$ mm for cases \#1 and \#2 respectively). However, slightly lower performances than the original pTV were achieved for the second 4DCT dataset \cite{Castillo2009a}. The study \cite{Castillo2009} shows smaller average displacements prior to image registration than the other two \cite{Castillo2009a,Castillo2013}, which may explain the variations in performances. Nevertheless, we  recall that these results were not optimized for VFS and that the only change made was to provide EBF components as substitutes for intensity images.

\begin{table}[htbp]
\centering
\caption{\label{tab1}DIR-Lab-4DCT datasets \cite{Castillo2009,Castillo2009a}. Comparison with two-step \textit{elastix} intensity registration. Mean TRE (in mm) for 300 landmarks.}
\begin{center}
\begin{tabular}{c|c c c c c }
\hline
\textbf{Subject} & {Orig.} & {ELX-1} & {ELX-2} & {VFS-1} &  {VFS-2}\\
\hline
Study \cite{Castillo2009a}\\
\#1 & $3.89$ & $1.29$ & $1.07$ & $\mathbf{1.06}$ & $\mathbf{1.06}$\\
\#2 & $4.34$ & $1.55$ & $1.08$ & $\mathbf{1.02}$ & ${1.05}$\\
\#3 & $6.94$ & $2.37$ & $1.56$ & ${1.44}$ & $\mathbf{1.40}$\\
\#4 & $9.83$ & $2.80$ & $2.00$ & $\mathbf{1.86}$ & ${1.88}$\\
\#5 & $7.48$ & $3.25$ & $2.26$ & $\mathbf{2.07}$ & $\mathbf{1.98}$\\
\hline
Study \cite{Castillo2009}\\
\#6 & $10.89$ & $3.60$ & $2.45$ & $2.87$ & $\mathbf{2.19}$\\
\#7 & $11.03$ & $5.04$ & $2.59$ & $3.49$ & $\mathbf{2.00}$\\
\#8 & $14.99$ & $7.40$ & $4.73$ & $4.33$ & $\mathbf{3.29}$\\
\#9 & $7.92$ & $3.00$ & $1.93$ & $1.98$ & $\mathbf{1.67}$\\
\#10 & $7.30$ & $3.06$ & $2.05$ & $2.20$ & $\mathbf{1.86}$\\
\hline
\end{tabular}
\end{center}
\end{table}

\begin{table}[htbp]
\centering
\caption{\label{tab2}DIR-Lab-COPd dataset \cite{Castillo2013}. Comparison with two-step \textit{elastix} intensity registration. Mean TRE (in mm) for 300 landmarks.}
\begin{center}
\begin{tabular}{c|c c c c c }
\hline
\textbf{Subject} & {Orig.} & {ELX-2} &  {VFS-2}\\
\hline
Study \cite{Castillo2013}\\
\#1 & $26.33$ & $13.59$ & $\mathbf{8.12}$ \\
\#2 & $21.79$ & $17.83$ & $\mathbf{6.74}$ \\
\#3 & $12.64$ & $5.40$ & $\mathbf{2.87}$ \\
\#4 & $29.58$ & $13.71$ & $\mathbf{11.03}$ \\
\#5 & $30.08$ & $14.20$ & $\mathbf{10.74}$ \\
\#6 & $28.46$ & $12.66$ & $\mathbf{7.07}$ \\
\#7 & $21.60$ & $7.26$ & $\mathbf{5.30}$ \\
\#8 & $26.46$ & $10.50$ & $\mathbf{8.02}$ \\
\#9 & $14.86$ & $6.43$ & $\mathbf{4.07}$ \\
\#10 & $21.81$ & $13.38$ & $\mathbf{10.75}$ \\
\hline
\end{tabular}
\end{center}
\end{table}

\begin{table}[htbp]
\centering
\caption{\label{tab3}DIR-Lab-4DCT datasets. Comparison with pTV intensity registration. mean TRE (in mm) for 300 landmarks}\begin{center}
\begin{tabular}{c|c c c c c }
\hline
\textbf{Subject} & orig. & {pTV-1} & {pTV-2} & {VFS}\\
\hline
Study \cite{Castillo2009a}\\
\#1 & $3.89$ &  $0.76$ & $0.77$ & $\mathbf{0.69}$ \\
\#2 & $4.34$& $0.77$ & $0.75$ & $\mathbf{0.67}$ \\
\#3 & $6.94$ & $0.90$ & $0.93$ & $\mathbf{0.87}$ \\
\#4 & $9.83$ & $1.24$ & $1.26$ & $\mathbf{1.22}$\\
\#5 & $7.48$ & $1.12$ & $\mathbf{1.07}$ & $1.11$\\
\hline
Study \cite{Castillo2009}\\
\#6 & $10.89$ & $0.85$& $\mathbf{0.83}$ &$0.95$\\
\#7 & $11.03$ & $\mathbf{0.80}$ & $\mathbf{0.80}$ & $0.87$\\
\#8 & $14.99$ & $1.34$ & $\mathbf{1.01}$ & $1.05$\\
\#9 & $7.92$ & $0.92$ & $\mathbf{0.91}$ &  $0.98$\\
\#10 & $7.30$ & $\mathbf{0.82}$ & $0.84$ & $0.88$\\
\hline
\end{tabular}
\end{center}
\end{table}

\subsection{\label{sec4b} Subcortical volume overlaps in MRI}  

\begin{figure}[htbp]
\subfigure[Hammers dataset]{\includegraphics[width=.95\linewidth]{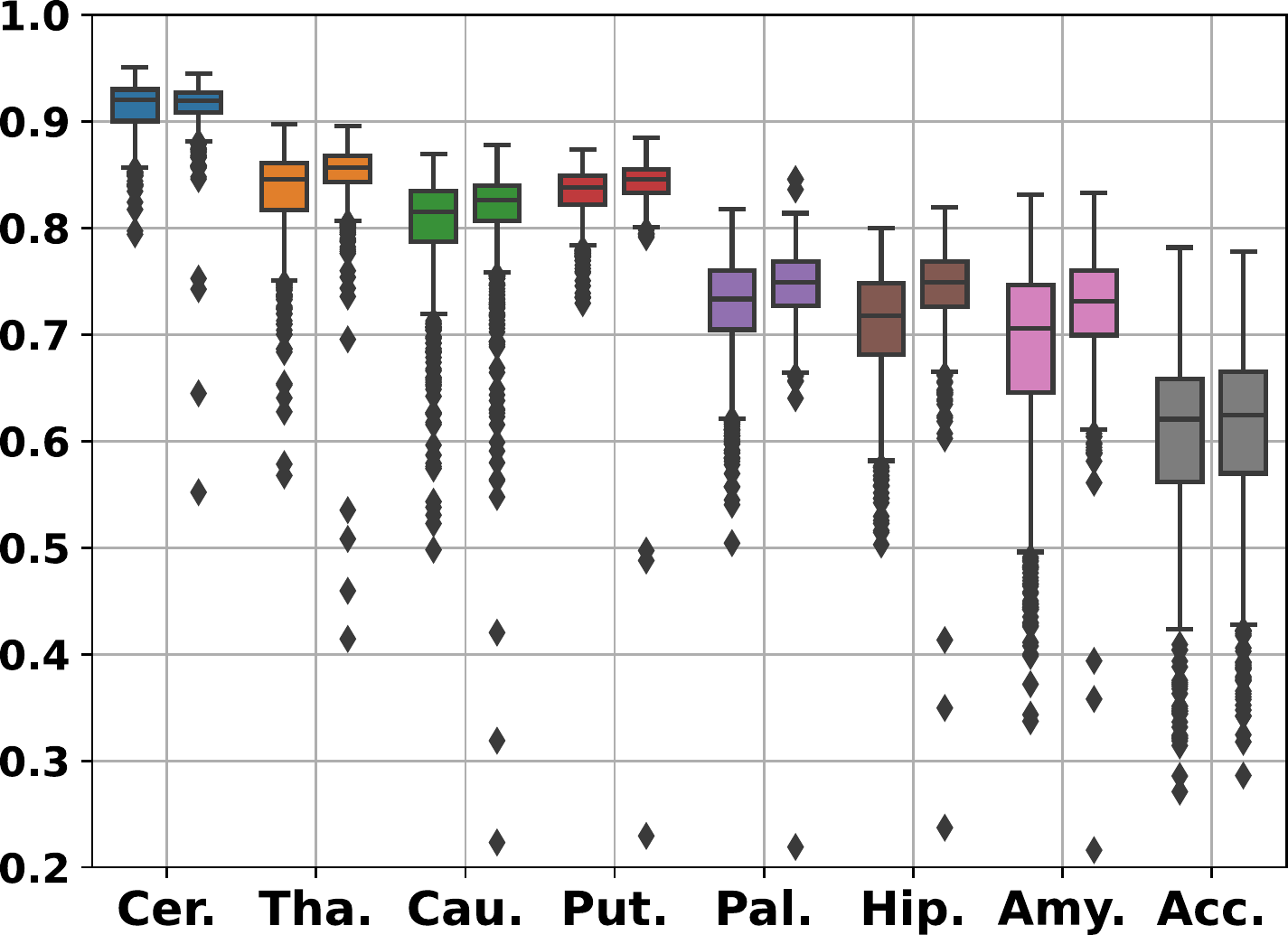}}
\subfigure[IBSR18 dataset]{\includegraphics[width=.95\linewidth]{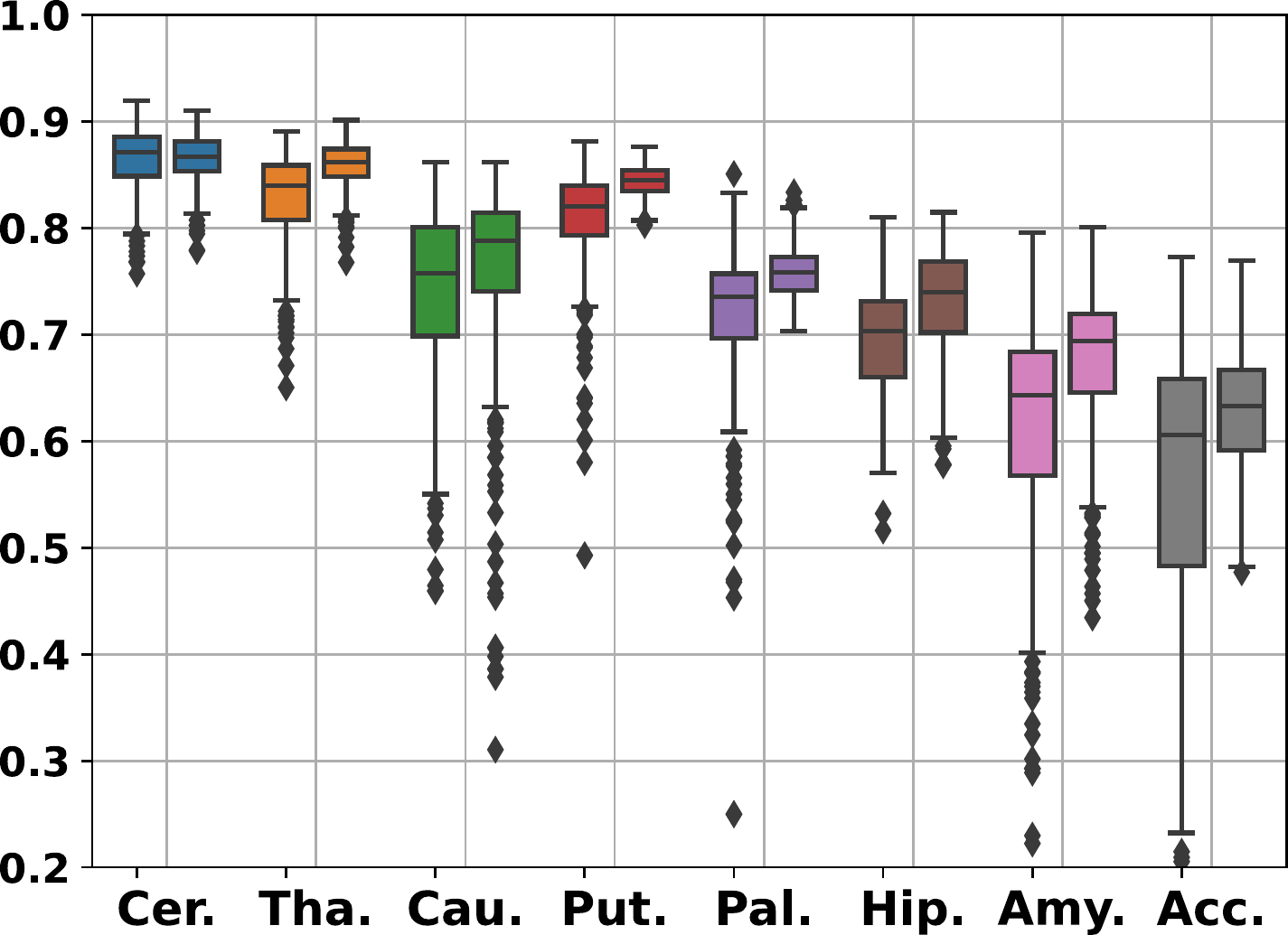}
}
\caption{\label{figBox} Box plots for tissue overlap scores as measured by the DSC in the (a) Hammers and (b) IBSR18 datasets in $8$ subcortical structures. The middle line represents the median. Results are shown as colored pairs in each structure for intensity-based (left box) and VFS-based (right box) registration. }
\end{figure}

\begin{table*}
\begin{center}
\caption{\label{tab4}Hammers dataset - DSC after registration (in \%) averaged over $8$ labeled subcortical regions for each subject. Standard deviation is between brackets.}
\begin{tabular}{l|c c c c c c c c c c c c|}
\hline
 \textbf{Method} &\#1&\#2&\#3&\#4&\#5&\#6&\#7&\#8&\#9&\#10 \\
\hline
\textbf{Qiao et al. \cite{Qiao2015} }& $76.8$ & $74.9$ & $76.9$ & $75.7$ & $77.2$ & $78.0$ & $74.4$ & $76.3$ & $71.6$ & $73.8$ \\
& ($9.5$) & ($13.8$) & ($11.8$) & ($12.8$) & ($10.3$) & ($10.3$) & ($14.9$) & ($12.6$) & ($15.2$) & ($13.2$) \\
\textbf{VFS} & $\mathbf{77.8}$ & $\mathbf{76.5}$ & $\mathbf{78.4}$ & $\mathbf{78.4}$ & $\mathbf{78.5}$ & $\mathbf{79.3}$ & $\mathbf{74.9}$ & $\mathbf{77.8}$ & $\mathbf{76.0}$ & $\mathbf{75.6}$ \\
& ($8.9$) & ($13.8$) & ($10.6$) & ($10.2$) & ($9.9$) & ($8.5$) & ($14.6$) & ($10.6$) & ($12.1$) & ($13.2$) \\
\hline
   &\#11&\#12&\#13&\#14&\#15&\#16&\#17&\#18&\#19&\#20  \\
 \hline

\textbf{Qiao et al. \cite{Qiao2015} }& $74.3$ & $76.5$ & $75.3$ & $77.2$ & $75.3$ & $75.2$ & $72.3$ & $76.1$ & $76.1$ & $76.6$ \\
 & ($12.5$) & ($13.0$) & ($12.1$) & ($10.9$) & ($13.3$) & ($12.0$) & ($15.7$) & ($11.0$) & ($11.0$) & ($12.0$) \\
\textbf{VFS}  & $\mathbf{78.6}$ & $\mathbf{77.7}$ & $\mathbf{78.7}$ & $\mathbf{77.4}$ & $\mathbf{76.6}$ & $\mathbf{78.1}$ & $\mathbf{76.1}$ & $\mathbf{76.7}$ & $\mathbf{79.2}$ & $\mathbf{77.6}$ \\
& ($9.0$) & ($13.8$) & ($9.1$) & ($11.4$) & ($13.5$) & ($10.3$) & ($11.7$) & ($10.0$) & ($8.8$) & ($12.6$) \\ 
\hline
   &\#21&\#22&\#23&\#24&\#25&\#26&\#27&\#28&\#29&\#30 \\
\hline
\textbf{Qiao et al. \cite{Qiao2015} } & $81.0$ & $79.6$ & $78.5$ & $79.1$ & $79.2$ & $76.9$ & $78.9$ & $80.9$ & $75.4$ & $76.5$ \\
& ($8.8$) & ($8.9$) & ($8.5$) & ($10.5$) & ($9.6$) & ($13.4$) & ($11.3$) & ($10.0$) & ($11.2$) & ($12.2$) \\
\textbf{VFS}  & $\mathbf{81.8}$ & $\mathbf{81.2}$ & $\mathbf{80.5}$ & $\mathbf{81.8}$ & $\mathbf{79.8}$ & $\mathbf{78.4}$ & $\mathbf{80.0}$ & $\mathbf{82.7}$ & $\mathbf{79.9}$ & $\mathbf{78.8}$ \\
& ($6.9$) & ($7.4$) & ($7.2$) & ($7.8$) & ($9.8$) & ($11.7$) & ($10.3$) & ($7.6$) & ($7.2$) & ($8.3$) \\ 

\hline
\end{tabular}
\end{center}
\end{table*}

\begin{table*}[htbp]
\begin{center}
\caption{\label{tab5}IBSR18 dataset - DSC after registration (in \%)  averaged over $8$ labeled subcortical regions for each subject. Standard deviation is between brackets.}
\begin{tabular}{l|c c c c c c c c c c c c|}
\hline
 \textbf{Method} & \#1 & \#2 &\#3&\#4&\#5&\#6&\#7&\#8&\#9 \\
\hline
\textbf{Qiao et al. \cite{Qiao2015} }& $75.0$ & $75.8$ & $69.9$ & $72.5$ & $74.0$ & $73.4$ & $75.7$ & $71.7$ & $75.6$  \\
& ($7.8$) & ($6.8$) & ($12.0$) & ($8.9$) & ($9.0$) & ($6.9$) & ($7.3$) & ($8.6$) & ($6.9$)\\
\textbf{VFS} &$\mathbf{76.8}$ & $\mathbf{77.6}$ & $\mathbf{73.8}$ & $\mathbf{75.6}$ & $\mathbf{75.2}$ & $\mathbf{76.2}$ & $\mathbf{76.4}$ & $\mathbf{75.7}$ & $\mathbf{77.4}$ \\
& ($9.6$) & ($8.2$) & ($11.6$) & ($10.4$) & ($10.1$) & ($7.7$) & ($9.8$) & ($9.2$) & ($9.3$)  \\
\hline
   &\#10&\#11&\#12&\#13&\#14&\#15&\#16&\#17&\#18\\
 \hline

\textbf{Qiao et al. \cite{Qiao2015}} & $68.7$ & $71.8$ & $73.0$ & $63.5$ & $75.1$ & $67.7$ & $76.9$ & $75.6$ & $79.0$ \\
 &  ($11.7$) & ($8.8$) & ($10.1$) & ($18.2$) & ($7.6$) & ($12.5$) & ($9.3$) & ($8.5$) & ($7.7$) \\
\textbf{VFS}  & $\mathbf{73.0}$ & $\mathbf{74.8}$ & $\mathbf{73.9}$ & $\mathbf{73.7}$ & $\mathbf{78.2}$ & $\mathbf{76.3}$ & $\mathbf{78.4}$ & $\mathbf{80.1}$ & $\mathbf{80.8}$\\
&  ($14.0$) & ($11.1$) & ($10.6$) & ($12.0$) & ($8.6$) & ($9.6$) & ($9.2$) & ($7.1$) & ($5.9$) \\ 
\hline
\end{tabular}
\end{center}
\end{table*}

 Table \ref{tab4} summarizes the scores obtained on the $30$ images, where a consistent improvement in volume overlap was achieved for all $30$ subjects. Fig. \ref{figBox} shows average DSC achieved in each region for the Hammers and the IBSR18 datasets. In the Hammers dataset, tissue overlap was improved across all regions using VFS, with most relative increase being achieved in the amygdala ($+3.97\%)$, hippocampus ($+3.46\%$)  pallidum ($+1.83\%$) and thalamus ($+1.78\%$) (Fig. \ref{figBox}a). DSC improvements with VFS were more pronounced on the IBSR18 dataset with $+8.44\%$ in the nuclei accumbens, $+6.07\%$ in the amygdala, $+4.16\%$ in the pallidum, $+3.61\%$ in the putamen,  $+3.59\%$ in the hippocampus and $+3.13\%$ in the thalamus (Fig. \ref{figBox}b). A reduction of the variability of the results was also observed for all subcortical regions. A likely explanation for the higher gains for VFS on the IBSR18 dataset is that the compared method by Qiao et al. \cite{Qiao2015} was optimized on the Hammers dataset.

After averaging scores over all $870$ registrations and all labels, a global DSC of $78.5\%\pm10.0\%$ was achieved for VFS against $76.5\%\pm11.4\%$ for the baseline. The $p$-value of the corresponding Wilcoxon signed rank test was $p=0.012$, suggesting statistical significance. 
 Table \ref{tab5} shows similar improvement of volumetric overlap scores on the IBSR18 dataset.

\begin{figure*}[htbp]
\includegraphics[width=\linewidth]{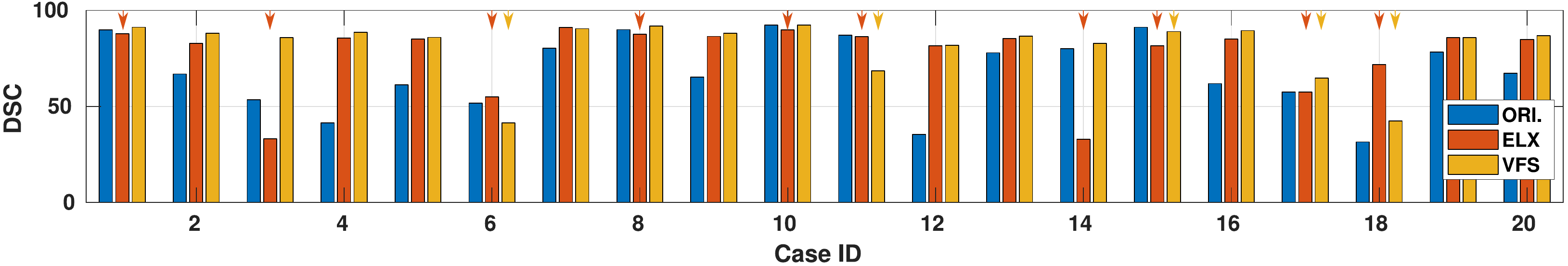}
\caption{\label{figChaos1} Results of renal volume overlaps using the pipeline proposed in \cite{Jansen2019} for the $20$ cases of the CHAOS dataset. Results marked using a red or yellow arrow are failure cases for ELX or VFS respectively, characterized by a decrease of DSC after registration or a DSC inferior to $80\%$.}
\end{figure*}
\subsection{\label{sec4c}Renal volume overlaps in multiparametric abdominal MRI}

Fig. \ref{figChaos1} shows renal volume overlaps for the $20$ images of the CHAOS dataset after registration with ELX or VFS.  Despite the similarity in registration scenarios between \cite{Jansen2019} and the present study, ELX led to several failure cases that we identified as a decrease of renal overlap measurements after registration or a DSC value inferior or equal to $80\%$ ($10$ cases out of $20$). Using the same pipeline, VFS led to only $5$ failure cases. Average overlap DSC was $68.1\%\pm18.8\%$ before image registration and $76.9\%\pm17.8\%$ and $81.1\%\pm15.13\%$ for ELX and VFS respectively, suggesting again better registration accuracy achieved with the substitution of  EBF-based structural representations for intensity images.

\section{Discussion and Conclusion}

We have proposed the use of new directional representations for image registration based on the similarity between regularized vector fields normally used in active contour segmentation, a technique we called vector field similarity. Results on a variety of registration scenarios (mono-modal inter-patient, multi-modal intra-patient) and anatomical locations show the potential advantage of such representations over the use of intensity images, with consistent improvement achieved through this substitution. As we adopt the point of view of structural representations, similarity can be measured using several distance metrics, both mono- and multimodal and be readily implemented and adapted into existing registration pipelines.  The main disadvantage associated to vector-valued directional representations over scalar $n-$dimensional intensity images is that they are $n$ times more memory consuming, which translates into longer registration times. However, we observed that among conventional metrics, VFS guided by the NCC similarity metric achieved on average results superior to, or on par with NMI even for pipelines optimized for NMI (not shown in the paper). The less memory-demanding NCC could therefore be substituted to NMI for VFS-based pipelines, partly compensating this drawback for various pipelines relying on NMI.
 
 VFS was evaluated within existing registration pipelines to provide a fair and objective comparison with current state-of-the-art. On the other hand, VFS could likely benefit from additional parameter tuning. 
Regarding parameter optimization, we used VFC-based similarity that enabled us to extend the capture range while providing a good localization of the global minimum for all values of the smoothing parameter $\gamma$.

Due to the simplicity of such a substitution, we hope this study will encourage the use of directional structural representations in cases where intensity-based registration does not seem to provide sufficient accuracy. Future investigations will be oriented towards the combination of directional structural representations with unsupervised deep learning-based registration.

\bibliographystyle{IEEEbib}

\begin{thebibliography}{00}

\bibitem{Ardekani2005}
B.~A. Ardekani, S. Guckemus, A.ABachman, M.JAHoptman, M.~
  Wojtaszek, and J.~Nierenberg,
\newblock ``Quantitative comparison of algorithms for inter-subject
  registration of 3D volumetric brain MRI scans,''
\newblock {\em J. Neurosci. Methods}, vol. 142, no. 1, pp. 67--76,
  2005.

\bibitem{Brock2006}
K.~K. Brock, L.~A. Dawson, M.~B. Sharpe, D.~J. Moseley, and
  D.~A. Jaffray,
\newblock ``Feasibility of a novel deformable image registration technique to
  facilitate classification, targeting, and monitoring of tumor and normal
  tissue,''
\newblock {\em Int. J. Radiat. Oncol.},
  vol. 64, no. 4, pp. 1245--1254, 2006.

\bibitem{Heinrich2013}
M.~P. Heinrich, M.~J., B.~W. Papie{\.z}, M. Brady,
  and J.~A. Schnabel,
\newblock ``Towards realtime multimodal fusion for image-guided interventions
  using self-similarities,''
\newblock in {\em MICCAI}. Springer, 2013, pp. 187--194.

\bibitem{Lorenzen2006}
P. Lorenzen, M. Prastawa, B. Davis, G. Gerig, E. Bullitt,
  and Sarang Joshi,
\newblock ``Multi-modal image set registration and atlas formation,''
\newblock {\em Med. Image Anal.}, vol. 10, no. 3, pp. 440--451, 2006.

\bibitem{Florkow2019}
M. C.~Florkow et~al.,
\newblock ``{The impact of MRI-CT registration errors on deep learning-based
  synthetic CT generation},''
\newblock in {\em Medical Imaging 2019: Image Processing}. SPIE, 2019, vol. 10949, p. 1094938.

\bibitem{Fayad2017}
H. Fayad, H. Schmidt, T. K{\"u}stner, and D. Visvikis,
\newblock ``{4-dimensional MRI and attenuation map generation in PET/MRI with
  4-dimensional PET-derived deformation matrices: study of feasibility for lung
  cancer applications},''
\newblock {\em {J. Nucl. Med.}}, vol. 58, no. 5, pp. 833--839,
  2017.

\bibitem{Zacharaki2009}
E.~I. Zacharaki, C.~S. Hogea, D. Shen, G. Biros, and
  C. Davatzikos,
\newblock ``Non-diffeomorphic registration of brain tumor images by simulating
  tissue loss and tumor growth,''
\newblock {\em Neuroimage}, vol. 46, no. 3, pp. 762--774, 2009.

\bibitem{Avants2014}
B.~B. Avants, N.~J. Tustison, M. Stauffer, G. Song, B. Wu,
  and J.~C. Gee,
\newblock ``The insight toolkit image registration framework,''
\newblock {\em Front. Neuroinform.}, vol. 8, pp. 44, 2014.

\bibitem{Klein2010}
Stefan K., M. Staring, K. Murphy, M.~A. Viergever, and J.~P.~W.
  Pluim,
\newblock ``{elastix: A Toolbox for Intensity-Based Medical Image
  Registration},''
\newblock {\em {IEEE Trans. Med. Imaging}}, vol. 29, pp. 196--205,
  2010.

\bibitem{Viergever2016}
M.~A. Viergever, J.~B.~A. Maintz, S. Klein, K. Murphy, M.
  Staring, and J.~P.~W. Pluim,
\newblock ``A survey of medical image registration – under review,''
\newblock {\em Med. Image Anal.}, vol. 33, pp. 140 -- 144, 2016,
\newblock 20th anniversary of the Med. Image Anal. journal (MedIA).

\bibitem{Sotiras2013}
A. Sotiras, C. Davatzikos, and N. Paragios,
\newblock ``Deformable medical image registration: A survey,''
\newblock {\em {IEEE Trans. Med. Imaging}}, vol. 32, no. 7, pp.
  1153--1190, 2013.

\bibitem{Klein2009}
S. Klein, J.~P.~W Pluim, M. Staring, and M.~A. Viergever,
\newblock ``{Adaptive stochastic gradient descent optimisation for image
  registration},''
\newblock {\em Int. J. Comput. Vis.}, vol. 81, no. 3, pp.
  227, 2009.

\bibitem{Vishnevskiy2016}
V. Vishnevskiy, T. Gass, G. Szekely, C. Tanner, and O. Goksel,
\newblock ``Isotropic total variation regularization of displacements in
  parametric image registration,''
\newblock {\em {IEEE Trans. Med. Imaging}}, vol. 36, no. 2, pp.
  385--395, 2016.

\bibitem{Viola1997}
P. Viola and W.~M. Wells~III,
\newblock ``Alignment by maximization of mutual information,''
\newblock {\em {Int. J. Comput. Vis.}}, vol. 24, no. 2, pp.
  137--154, 1997.

\bibitem{Studholme1999}
C. Studholme, D.~LG Hill, and D.~J. Hawkes,
\newblock ``{An overlap invariant entropy measure of 3D medical image
  alignment},''
\newblock {\em {Pattern Recognit.}}, vol. 32, no. 1, pp. 71--86, 1999.

\bibitem{Haber2006}
E. Haber and J. Modersitzki,
\newblock ``Intensity gradient based registration and fusion of multi-modal
  images,''
\newblock in {\em MICCAI}. Springer, 2006, pp. 726--733.

\bibitem{Heinrich2012}
M.~P. Heinrich et~al.,
\newblock ``{MIND: Modality independent neighbourhood descriptor for
  multi-modal deformable registration},''
\newblock {\em {Med. Image Anal.}}, vol. 16, no. 7, pp. 1423--1435, 2012.

\bibitem{Loeckx2009}
D. Loeckx, P. Slagmolen, F. Maes, D. Vandermeulen, and P.
  Suetens,
\newblock ``Nonrigid image registration using conditional mutual information,''
\newblock {\em {IEEE Trans. Med. Imaging}}, vol. 29, no. 1, pp.
  19--29, 2009.

\bibitem{Simonovsky2016}
M. Simonovsky, B. Guti{\'e}rrez-Becker, D. Mateus, N.
  Navab, and N. Komodakis,
\newblock ``A deep metric for multimodal registration,''
\newblock in {\em MICCAI}. Springer, 2016, pp. 10--18.

\bibitem{Pluim2000}
J.~P.~W. Pluim, J.~B.~A. Maintz, and M.~A. Viergever,
\newblock ``Image registration by maximization of combined mutual information
  and gradient information,''
\newblock in {\em MICCAI}. Springer, 2000, pp. 452--461.

\bibitem{Murphy2011}
K. Murphy et~al.,
\newblock ``{Evaluation of registration methods on thoracic CT: the EMPIRE10
  challenge},''
\newblock {\em {IEEE Trans. Med. Imaging}}, vol. 30, no. 11, pp.
  1901--1920, 2011.

\bibitem{Blendowski2020}
M. Blendowski, L. Hansen, and M.~P. Heinrich,
\newblock ``Weakly-supervised learning of multi-modal features for regularised
  iterative descent in 3D image registration,''
\newblock {\em Med. Image Anal.}, p. 101822, 2020.

\bibitem{Devos2019}
B. {de Vos}, F. Berendsen, M. A. Viergever, H. Sokooti, M.
  Staring, and I. Išgum,
\newblock ``A deep learning framework for unsupervised affine and deformable
  image registration,''
\newblock {\em Med. Image Anal.}, vol. 52, pp. 128 -- 143, 2019.

\bibitem{Ronneberger2015}
O. Ronneberger, P. Fischer, and T. Brox,
\newblock ``U-net: Convolutional networks for biomedical image segmentation,''
\newblock in {\em MICCAI}. Springer, 2015, pp. 234--241.

\bibitem{Wachinger2012}
C. Wachinger and N. Navab,
\newblock ``Entropy and laplacian images: Structural representations for
  multi-modal registration,''
\newblock {\em {Med. Image Anal.}}, vol. 16, no. 1, pp. 1--17, 2012.

\bibitem{Lee2019}
M.~C.~H. Lee, O. Oktay, A. Schuh, M. Schaap, and B. Glocker,
\newblock ``Image-and-spatial transformer networks for structure-guided image
  registration,''
\newblock in {\em MICCAI}. Springer, 2019, pp. 337--345.

\bibitem{Arar2020}
M. Arar, Y. Ginger, D. Danon, A.~H. Bermano, and D. Cohen-Or,
\newblock ``Unsupervised multi-modal image registration via geometry preserving
  image-to-image translation,''
\newblock in {\em IEEE CVPR}, 2020, pp. 13410--13419.

\bibitem{Jaderberg2015}
M. Jaderberg, K. Simonyan, and A. Zisserman,
\newblock ``Spatial transformer networks,''
\newblock in {\em NeurIPS}, 2015, pp.
  2017--2025.

\bibitem{Xu1998}
C. Xu and J.~L. Prince,
\newblock ``Snakes, shapes, and gradient vector flow,''
\newblock {\em {IEEE Trans. Image Process.}}, vol. 7, no. 3, pp.
  359--369, 1998.

\bibitem{Li2007}
B. Li and S.~T. Acton,
\newblock ``Active contour external force using vector field convolution for
  image segmentation,''
\newblock {\em {IEEE Trans. Image Process.}}, vol. 16, no. 8, pp.
  2096--2106, 2007.

\bibitem{Jaouen2014}
V. Jaouen et~al.,
\newblock ``{Variational segmentation of vector-valued images with gradient
  vector flow},''
\newblock {\em {IEEE Trans. Image Process.}}, vol. 23, no. 11, pp.
  4773--4785, 2014.

\bibitem{Jaouen2019}
V. {Jaouen}, J. {Bert}, N. {Boussion}, H. {Fayad}, M. {Hatt}, and D. {Visvikis},
\newblock ``{Image Enhancement With PDEs and Nonconservative Advection Flow
  Fields},''
\newblock {\em {IEEE Trans. Image Process.}}, vol. 28, no. 6, pp.
  3075--3088, 2019.

\bibitem{Jaouen2019a}
V. {Jaouen} et~al.,
\newblock ``{Prostate Volume Segmentation in TRUS Using Hybrid
  Edge-Bhattacharyya Active Surfaces},''
\newblock {\em {IEEE. Trans. Biomed. Eng.}}, vol. 66, no. 4,
  pp. 920--933, 2019.

\bibitem{Xu2020}
C. Xu and J.~L. Prince,
\newblock {\em Gradient Vector Flow}, pp. 1--8,
\newblock Springer International Publishing, Cham, 2020.

\bibitem{Bazan2007}
C. Bazan and P. Blomgren,
\newblock ``Image smoothing and edge detection by nonlinear diffusion and
  bilateral filter,''
\newblock {\em Research Report CSRCR 2007-21, San Diego State University}, vol. 21, pp. 2--15, 2007.

\bibitem{Cocosco1997}
C.~A. Cocosco, V. Kollokian, R.~K.~S. Kwan, G.~B. Pike, and A.~C.
 Evans,
\newblock ``{BrainWeb: Online interface to a 3D MRI simulated brain
  database},''
\newblock in {\em Neuroimage}. Citeseer, 1997.

\bibitem{Xu1998a}
C. Xu and J.~L. Prince,
\newblock ``Generalized gradient vector flow external forces for active
  contours,''
\newblock {\em Signal process.}, vol. 71, no. 2, pp. 131--139, 1998.

\bibitem{Jaouen2013}
V. Jaouen, P. Gonzalez, S. Stute, D. Guilloteau, I. Buvat,
  and C. Tauber,
\newblock ``{Vector-based active surfaces for segmentation of dynamic PET
  images},''
\newblock in {\em {IEEE ISBI}}. IEEE, 2013, pp. 61--64.

\bibitem{Rohlfing2013}
T. Rohlfing,
\newblock ``Image similarity and tissue overlaps as surrogates for image
  registration accuracy: widely used but unreliable,''
\newblock {\em IEEE Trans. Med. Imaging}, vol. 31, no. 2, pp.
  153--163, 2011.

\bibitem{Castillo2009}
E. Castillo, R. Castillo, J. Martinez, M. Shenoy, and T.
  Guerrero,
\newblock ``Four-dimensional deformable image registration using trajectory
  modeling,''
\newblock {\em Phys. Med. Biol.}, vol. 55, no. 1, pp. 305, 2009.

\bibitem{Castillo2013}
R. Castillo et~al.,
\newblock ``{A reference dataset for deformable image registration spatial
  accuracy evaluation using the COPDgene study archive},''
\newblock {\em Phys. Med. Biol.}, vol. 58, no. 9, pp. 2861, 2013.

\bibitem{Modersitzki2009}
J. Modersitzki,
\newblock {\em {FAIR: Flexible Algorithms for Image Registration}},
\newblock SIAM, USA, 2009.

\bibitem{Tustison2014}
N.~J. Tustison et~al.,
\newblock ``{Large-scale evaluation of ANTs and FreeSurfer cortical thickness
  measurements},''
\newblock {\em {Neuroimage}}, vol. 99, pp. 166--179, 2014.

\bibitem{Hammers2003}
A. Hammers et~al.,
\newblock ``Three-dimensional maximum probability atlas of the human brain,
  with particular reference to the temporal lobe,''
\newblock {\em Hum. Brain Mapp.}, vol. 19, no. 4, pp. 224--247, 2003.

\bibitem{Chaos2020}
A.~E. Kavur et~al.,
\newblock ``{CHAOS Challenge - Combined (CT-MR) Healthy Abdominal Organ
  Segmentation},'' Jan. 2020.

\bibitem{Castillo2009a}
R. Castillo et~al .,
\newblock ``A framework for evaluation of deformable image registration spatial
  accuracy using large landmark point sets,''
\newblock {\em Phys. Med. Biol.}, vol. 54, no. 7, pp. 1849, 2009.

\bibitem{Qiao2015}
Y. Qiao, B. van Lew, B.~P.~F. Lelieveldt, and M. Staring,
\newblock ``Fast automatic step size estimation for gradient descent
  optimization of image registration,''
\newblock {\em IEEE Trans. Med. Imaging}, vol. 35, no. 2, pp.
  391--403, 2015.

\bibitem{Carrillo2000}
A. Carrillo, J.~L. Duerk, J.~S. Lewin, and D.~L. Wilson,
\newblock ``{Semiautomatic 3-D image registration as applied to interventional
  MRI liver cancer treatment},''
\newblock {\em {IEEE Trans. Med. Imaging}}, vol. 19, no. 3, pp.
  175--185, 2000.

\bibitem{Jansen2019}
M.~J~.A. Jansen et~al.,
\newblock ``Liver segmentation and metastases detection in MR images using
  convolutional neural networks,''
\newblock {\em J. Med. Imaging}, vol. 6, 044003, 2019.

\end{thebibliography}

\end{document}